\pdfoutput=1

\documentclass[11pt]{article}

\usepackage[final]{acl}

\usepackage{times}
\usepackage{latexsym}

\usepackage[T1]{fontenc}

\usepackage[utf8]{inputenc}

\usepackage{microtype}

\usepackage{inconsolata}

\usepackage{graphicx}
\usepackage{booktabs}
\usepackage{amsmath}
\usepackage{multirow}
\usepackage{lineno}
\usepackage{subcaption}
\title{Using Similarity to Evaluate Factual Consistency in Summaries}

\author{Yuxuan Ye \and Edwin Simpson \and Raul Santos Rodriguez \\
        Intelligent Systems Laboratory \\ University of Bristol \\
        \texttt{\{yuxuan.ye, edwin.simpson, enrsr\}@bristol.ac.uk}}

\begin{document}
\maketitle
\begin{abstract}
Cutting-edge abstractive summarisers generate fluent summaries, but the factuality of the generated text is not guaranteed. 
Early summary factuality evaluation metrics are usually based on n-gram overlap and embedding similarity, but are reported fail to align with human annotations.
Therefore, many techniques for detecting factual inconsistencies build pipelines around natural language inference (NLI) or question-answering (QA) models with additional supervised learning steps. 
In this paper, we revisit similarity-based metrics,
showing that this failure stems from the comparison text selection and its granularity. 
We propose a new zero-shot factuality evaluation metric,
Sentence-BERT Score (SBERTScore), which compares sentences between the summary and the source document. 
It outperforms widely-used word-word metrics including BERTScore and can compete with existing NLI and QA-based factuality metrics on the benchmark without needing any fine-tuning.
Our experiments indicate that each technique has different strengths, with SBERTScore particularly effective in identifying correct summaries.
We demonstrate how a combination of techniques is more effective in detecting various types of error.\footnote{The code will be made available upon acceptance.}
\end{abstract}

\section{Introduction}

The rapid development of natural language generation techniques has created new challenges for evaluation, since evaluation metrics have not undergone the same pace of improvement.
For instance, ROUGE \cite{lin-2004-rouge} has been involved in summary evaluation for decades and is still one of the most widely applied metrics for the overall quality of generated summaries \cite{koto-2022-ffci}, despite comparing only lexical, rather than semantic, overlap.
Abstractive summarisers have set new records for ROUGE scores many times in recent years \cite{zhang-2020-pegasus,lewis-2019-bart,zhao-2022-calibrating}, but research shows that they are prone to generate factually inconsistent summaries that cannot be reflected by ROUGE scores \cite{maynez-2020-faithfulness,pagnoni-2021-understanding,durmus-2020-feqa},
as ROUGE cannot distinguish between valid paraphrases and factual inconsistencies.

Recent factuality metrics fall into two types. 1) NLI-based metrics \cite{kryscinski-2020-evaluating,laban-2022-summac} predict the probability that each part of the given summary is entailed by the source document and combine these predictions to form an overall score. 
2) QA-based metrics \cite{durmus-2020-feqa,fabbri-2021-qafacteval,scialom-2021-questeval} simulate the process of a human performing reading comprehension tasks and compute the factuality score based on how many questions generated from the summary can be correctly answered from the given source document. 
These two paradigms need to train their models on a large-scale dataset, but existing factuality datasets are usually insufficient.
%

In this paper, we develop a metric that does not require additional training when applying it to a new domain by making use of pretrained sentence embeddings. 
Similarity-based metrics are proposed to handle synonyms that fail the n-gram-based methods \citep{zhang-2019-bertscore}. 
However, early exploration indicates that they still can't align well with human factuality annotations \citep{maynez-2020-faithfulness, pagnoni-2021-understanding,durmus-2020-feqa}.

Our work shows how BERTScore \citep{zhang-2019-bertscore} can provide useful factuality metrics if it is used to compare generated summaries to sources, which are the same inputs as NLI and QA-based metrics take, rather than reference summaries.
However, comparing individual words offers very limited insights into factual consistency, as sentences can be constructed in entirely different ways.
Therefore, we propose a sentence-level factuality evaluation metric, SBERTScore.
It computes cosine similarity between sentence embeddings \cite{reimers-2019-sbert}, which take all words in the sentence into consideration, including their order and composition,
so can better represent the semantics of the complete sentence compared to the contextualised word embeddings used by BERTScore.
Comparison on a factuality benchmark \citep{tang-2023-understanding} shows that SBERTScore outperforms the widely used token-level and n-gram-level metrics, BERTScore and ROUGE.

We also compare SBERTScore against recent NLI and QA-based factuality metrics. 
The experimental results show that SBERTScore can outperform NLI-based metrics in the same zero-shot setting and is competitive with QA-based metrics. 
In addition, SBERTScore does not require any additional training steps as it benefits from high-quality general-purpose pretrained embeddings, and has much less computational complexity at inference time.
Importantly, our design of SBERTScore avoids truncating long source documents, instead selecting an appropriate granularity to segment the sources before feeding them into the sentence transformer.
Further analysis of agreement between metrics, as well as the types of errors \cite{tang-2023-understanding} they detect, shows that SBERTScore can capture different kinds of errors than NLI and QA-based methods. 
We show that even a simple combination of metrics can outperform the individual base metrics, which suggests that combining diverse metrics may be a promising direction for future research.

Our contributions are three-fold:
\begin{itemize}
    \item We propose a zero-shot technique for evaluating the factual consistency of summaries using pretrained embeddings off-the-shelf.
    \item We conduct an empirical evaluation, which reveals that the previous underperformance of methods such as BERTScore is due to the use of reference summaries. SBERTScore is competitive with recent factuality metrics on the benchmark without requiring additional training steps as other metrics do.
    \item We show that different evaluation metrics are necessary to capture different types of error, and introduce a simple combination that outperforms the state of the art.
\end{itemize}

\section{Related Work}

\subsection{NLI-based Factuality Metrics} 
The NLI task is similar to predicting factual consistency between source document and generated summary. Hence, previous research \cite{barrantes-2020-adversarial,falke-2019-ranking} attempted to transfer models trained on NLI datasets to factual consistency detection. However, a subsequent study \cite{kryscinski-2020-evaluating} showed that those NLI models are only as good as random guessing. Therefore, a series of work \cite{kryscinski-2020-evaluating,laban-2022-summac} made efforts to build up datasets for training factuality metrics. Although the dataset can be synthesised using entity swap to save the effort of collecting human annotations, the error distribution is not the same as real summaries \cite{pagnoni-2021-understanding}.

Another strand of research into NLI-based factuality prediction focused on the granularity of the input text. Early works \cite{barrantes-2020-adversarial,falke-2019-ranking,kryscinski-2020-evaluating} concatenate the system summary with the whole source document as the input. Firstly, this often requires truncating the source document to fit the length limit, which can lead to underestimating factuality due to the information loss. Secondly, the NLI models applied in their work are trained on much shorter sentence pairs. Directly applying these models on long text such as source documents does not align with their training data distribution. Following work \cite{goyal-2020-evaluating,laban-2022-summac} investigated the effect of performing inference at different levels, including word, dependency, sentence, and paragraph, revealing that segmenting source documents into sentences and dependency arcs is more suitable for current NLI models.
This inspired us to explore the suitability of different input text granularities for similarity-based evaluation metrics, which have not been investigated in past work.

\subsection{QA-based Factuality Metrics} 
QA-based metrics \cite{chen-2018-semantic,wang-2020-asking,durmus-2020-feqa,fabbri-2021-qafacteval} assemble multiple modules with different functions. 
An answer selection module first selects a set of answers from the summary, usually including named entities and noun phrase chunks. A question generation module conditioned upon the selected answers is applied on the summary as context to raise questions. 
The QA component answers the generated questions conditioned on the given source document. 
The final score is then computed on the overlapping extent of the two answer sets. This paradigm provides an interpretable way to assess factuality by showing questions with inconsistent answers. 
However, since several text generation models are involved in the evaluation process, this methodology usually requires a large training dataset and is time-consuming at inference time. 
We were therefore motivated to investigate alternatives, 
as factuality datasets are usually small and domain-specific, and the evaluation process is expected to be prompt.

\subsection{Similarity-based Factuality Metrics}
BERTScore \cite{zhang-2019-bertscore} is used as a stronger baseline than ROUGE \cite{lin-2004-rouge} in factual consistency detection, but it does not correlate well with human judgements \cite{pagnoni-2021-understanding}. Koto et al. \shortcite{koto-2022-ffci} adapted BERTScore by averaging the three highest scores and showed that it can detect the information overlap of system summaries and source documents, but there was still a large performance gap with other metrics \cite{fabbri-2021-qafacteval}. Previous work only explored word-level similarity, while other paradigms work on coarser text pieces. This inspired us to investigate the performance of sentence-level similarity-based metrics.

\subsection{Evaluation for Factuality Metrics} 
Factuality metrics are usually evaluated using correlations between the metric scores and human annotations \cite{pagnoni-2021-understanding} or as binary classifiers that label summaries as consistent or inconsistent~\cite{tang-2023-understanding, laban-2022-summac,fabbri-2021-qafacteval, kryscinski-2020-evaluating}.
With more recent benchmarks, correlation has become a less well-suited metric since most of the human annotations are binary labels.
When evaluating metrics as binary classifiers, balanced accuracy is applied to eliminate the effect of imbalanced data distribution.
The threshold to split metric scores into binary labels, as well as any other hyperparameters, will be tuned on the validation set of the benckmark.
An alternative to balanced accuracy is Area Under Curve of Receiver Operating Characteristic (ROC-AUC) \cite{fawcett-2006-roc}, which measures the ability of the metric to discriminate consistent and inconsistent summaries without fixing a particular threshold. 

\section{Sentence-BERT Score}\label{sec:sbertscore}
BERTScore \cite{zhang-2019-bertscore} computes similarity t the word-level by comparing the embeddings of words in the generated text with their closest match in the source or reference text. 
However, factual consistency should be judged at a higher level as sentences containing similar words can express different meanings.
Therefore, we propose the sentence-level evaluation metric, Sentence-BERT Score (SBERTScore), utilising sentence transformers to capture the meaning of the complete sentence.
The \textit{precision} and \textit{recall} of our proposed metric are defined as follows. $S_{\{D,S\}}$ represent the sentence set of the given source document and summary respectively, and $s_{\{i,j\}}$ are the sentences in the sets.

\begin{equation*}
    SBERT_{prec}  = \frac{1}{|S_S|}\sum_{s_i \in S_S}{\max_{s_j \in S_D}cossim(s_i, s_j)}
\end{equation*}
\begin{equation*}
    SBERT_{recall}  = \frac{1}{|S_D|}\sum_{s_j \in S_D}{\max_{s_i \in S_S}cossim(s_i, s_j)}
\end{equation*}

In practice, sentence transformers \cite{reimers-2019-sbert} can generate embeddings for any texts shorter than 512 tokens, which need not be single, complete sentences.
Therefore, we investigate three different granularities, and test them in Section \ref{sec:granularity} to find the most suitable setup for SBERTScore:

\paragraph{Sent} Segment the input text into 
sentences.
\paragraph{Doc} Take the whole 
text as input and truncate
the part that exceeds the length limit.
\paragraph{Mean} Segment the input text into sentences 
and take the average sentence embedding to represent the whole input.

Considering \textit{precision}, \textit{recall} and \textit{F1 measure}: precision is better suited to capturing factuality because it reflects the  extent to which summary sentences are supported by source sentences. 
We test this hypothesis in the following Section \ref{sec:selection}. 

\subsection{Computational Efficiency}
SBERTScore applies an all-purpose embedding model as the backbone, which provides reliable sentence embeddings that can be used out-of-box without 
the cost of additional training, in contrast to other metrics based on NLI or QA.
SBERTScore also has advantages at inference time. 
We denote the number of sentences in the system summary and source document as $N$ and $M$ respectively.
The majority of inference time is spent on calling the backbone model to process the input sentences.
NLI-based metrics need to take each sentence pair once, therefore the number of inputs that the backbone model processes is $\mathcal{O}(NM)$. SBERTScore uses a similar backbone but only needs to compute the embedding once for each sentence, so the complexity is $\mathcal{O}(N+M)$. 
As for QA-based metrics, its runtime is much greater than the other two as multiple models are involved in question generation and answering, thus has the lowest efficiency. 
We randomly sampled 1000 pieces of data from the benchmark, and test the runtime of QuestEval \cite{scialom-2021-questeval}, SummaC$_{\{ZS, Conv\}}$ \cite{laban-2022-summac}, BERTScore \cite{zhang-2019-bertscore} and SBERTScore on Intel(R) Core(TM) i9-10900X CPU @ 3.70GHz with NVIDIA A5000.
Results in Appendix \ref{sec:speed} show that SBERTScore only comes after BERTScore in processing speed, and is 3 times faster than the rival NLI-based method SummaC$_{\{ZS, Conv\}}$ and 30 times faster than the QA-based metric QuestEval.

\section{Experimental Settings}

\subsection{Datasets}
To evaluate our proposed factuality metric against alternatives, we use the benchmark built by Tang et al. \shortcite{tang-2023-understanding}, which consists of summaries and human annotations sampled from nine existing factuality datasets, including XSumFaith (XSF) \cite{maynez-2020-faithfulness}, Polytope \cite{huang-2020-have}, FactCC \cite{kryscinski-2020-evaluating}, SummEval \cite{fabbri-2021-summeval}, FRANK \cite{pagnoni-2021-understanding}, QAGS \cite{wang-2020-asking}, CLIFF \cite{cao-2021-cliff}, Goyal 21' \cite{goyal-2021-annotating}, and XENT \cite{cao-2021-hallucinated}. The dataset characteristics are shown in Table \ref{tab:dataset}.
\begin{table}[htbp]
    \centering
    \small
    \begin{tabular}{lcccc}
    \toprule
    \multirow{2}{*}{Dataset} & Annotator & \multirow{2}{*}{Size} & Source & Summary \\
     & Number & & Length & Length \\
    \midrule
    XSF & 3 & 2353 & 505.0 & 28.1 \\
    Polytope & 3 & 1268 & 691.5 & 83.1 \\
    FactCC & 2 & 1434 & 728.4 & 21.8 \\
    SummEval & 8 & 1698 & 453.7 & 79.2 \\
    FRANK & 3 & 1393 & 692.1 & 67.5 \\
    QAGS & 3 & 474 & 414.2 & 45.9 \\
    CLIFF & 2 & 600 & 576.9 & 45.8 \\
    Goyal' 21 & 2 & 100 & 504.3 & 29.9 \\
    XENT & 5 & 696 & 436.6 & 32.9 \\
    \midrule
    Average & 3.4 & 1112.8 & 572.8 & 50.4 \\
    \bottomrule
    \end{tabular}
    \caption{Dataset characteristics in the benchmark. Source Length and Summary Length are the token numbers in the source and summary counted based on the results of Roberta-large tokenizer \citep{liu-2019-roberta}.}
    \label{tab:dataset}
\end{table}
All source documents are English news articles, originally from the validation and test set of two news summarisation benchmarks, CNNDM \cite{see-2017-cnndm} and XSum \cite{narayan-2018-xsum}. 
Corresponding summaries were generated by a range of abstractive summarisers, including BART \cite{lewis-2019-bart}, PEGASUS \cite{zhang-2020-pegasus}, and BERTSumAbs \cite{liu-2019-fine}. 
We remove data from CNNDM in Goyal 21', as its validation set is extremely imbalanced (only 1 consistent example in the validation set), which impairs the classification threshold selection. 

\subsection{Performance Evaluation}
Since the label distribution varies across datasets, 
we use balanced accuracy \cite{laban-2022-summac},
defined as:
\begin{equation*}
\label{eq:bl_acc}
    BalancedAcc = \frac{1}{2}\left(\frac{TP}{TP+FN}+\frac{TN}{TN+FP}\right),
\end{equation*}
where $TP$, $TN$, $FP$, and $FN$ refer to the number of true positives, true negatives, false positives, and false negatives, respectively. 
We select a threshold for each metric using the validation set to compute balanced accuracy. ROC-AUC is also reported to demonstrate the metric's ability to distinguish consistent and inconsistent summaries.

\subsection{Evaluation Metrics for Comparison}
\label{sec:comparison}
This section introduces factuality metrics studied for comparison.

\paragraph{QAFactEval} Fabbri et al. \shortcite{fabbri-2021-qafacteval} conducted a comprehensive evaluation of the components of QA-based metrics. They aggregated more advanced models into the system and optimised a pipeline for computing consistency scores.

\paragraph{QuestEval} Scialom et al. \shortcite{scialom-2021-questeval} proposed a QA-based framework to compute consistency scores for given text pairs. They first select an answer set from the candidate text, then generate questions using the other text as input with conditions from the answer set. The QA module answers the questions and the overlap between the two answer sets is counted to obtain precision and recall.
They use F1 measure as the final factual consistency score.

\paragraph{DAE} Goyal and Durrett \shortcite{goyal-2020-evaluating} extract dependencies from given texts using the parse tree. They train a model to predict entailment at the dependency-level. The final score is the average entailment score over all dependency arcs in the given source and summary.

\paragraph{SummaC$_{\{ZS,Conv\}}$} Laban et al. \shortcite{laban-2022-summac} train a sentence-level NLI model and compute the entailment scores for all pairs of sentences from the source document and the summary. \textbf{ZS} stands for zero-shot, where the final entailment score is the average of the maximum entailment score for each sentence in the summary. \textbf{Conv} is a variant with an extra learned convolutional layer that aggregates the entailment score matrix to a final score.

\paragraph{ROUGE} Lin \shortcite{lin-2004-rouge} propose an evaluation metric by counting the overlapping words between the given reference and candidate text pairs.

\paragraph{BERTScore} Zhang et al. \shortcite{zhang-2019-bertscore} report the average cosine similarity of the matched word embeddings provided by BERT \cite{devlin-2018-bert} or other related models.

FactCC, SummaC$_{\{ZS, Conv\}}$, DAE are NLI-based metrics, and QuestEval, QAFactEval are QA-based metrics. To have a fair comparison, we use the pretrained RoBERTa-large \cite{liu-2019-roberta} as the backbone for BERTScore and all-roberta-large-v1 \cite{reimers-2019-sbert} for SBERTScore. The two checkpoints have identical numbers of layers, and the only difference is that they are trained for different text embeddings.

\section{Experiments and Results}

In this section, we first investigate the suitability of different settings for similarity-based metrics. 
We also look into a case study to better understand the metrics' behaviour when processing negation and neutral sentences. 
Then we test metric performance on the benchmark. 
The last subsection reports the error analysis and agreement between different factuality metrics and demonstrates the benefit of metric combination.

\subsection{Comparison of Precision, Recall, and F1}
\label{sec:selection}
We compare \textit{precision}, \textit{recall}, and \textit{F1 measure} to select the most informative measure for similarity-based metrics. From the definition, \textit{precision} relates better to the accuracy of the information included in the summary, while \textit{recall} reflects how completely the summary covers the source document. 
Table \ref{tab:precision} supports our hypothesis that \textit{precision} can assess generated summaries more accurately from the perspective of factuality. Therefore, we report \textit{precision} of BERTScore and SBERTScore in the following sections.

\subsection{Comparison Text Selection}
We investigate the effect of taking \texttt{(source, summary)} and \texttt{(reference, summary)} as input to n-gram matching and similarity-based metrics. 
Table \ref{tab:source} shows that the choice of comparison text makes a huge difference to the same evaluation metric. 
The highest results on \texttt{(reference, summary)} pairs are only as good as a random guess, while the performance on \texttt{(source, summary)} pairs is greatly improved. 
References may be unsuitable since
they carry less information than the source document, and often contain extrinsic knowledge aggregated by human writers \cite{maynez-2020-faithfulness}, especially in XSum \cite{narayan-2018-xsum}.

\begin{table}[htbp]
    \centering
    \small
    \begin{tabular}{cccc}
    \toprule
     Measure & Precision & Recall & F1 \\
    \midrule
     BERTScore & 0.758 & 0.627 & 0.710 \\
     SBERTScore & \textbf{0.779} & 0.644 & 0.703 \\
    \bottomrule
    \end{tabular}
    \caption{Average balanced accuracy on the benchmark using \textit{precision}, \textit{recall}, and \textit{F1 measure}. The highest result is in \textbf{bold}, which is significantly higher than the second best result with $p<0.05$.}
    \label{tab:precision}
\end{table}

\begin{table}[htbp]
    \centering
    \small
    \begin{tabular}{cccc}
    \toprule
     Metric & Reference & Source \\
    \midrule
     Rouge 1 & 0.491 & 0.638 \\
     Rouge 2 & 0.318 & 0.706 \\
     Rouge L & 0.491 & 0.674 \\
     BERTScore & 0.500 & 0.759 \\
     SBERTScore & 0.499 & \textbf{0.779} \\
    \bottomrule
    \end{tabular}
    \caption{Average balanced accuracy (Balanced Acc.) computed on different comparison texts on the benchmark. All results in the source column are significantly higher than their corresponding results in the upper bracket with $p<0.05$.}
    \label{tab:source}
\end{table}

\subsection{Text Granularity Selection}
\label{sec:granularity}
As performance can vary based on how the input text is segmented and processed before being fed into the sentence-transformer, we test the settings mentioned above in different combinations to build up a recommendation for using SBERTScore.
\begin{table}[bhtp]
    \centering
    \small
    \begin{tabular}{clc}
    \toprule
    Model & Granularity & Balanced Accuracy\\
    \midrule
    BERTScore & Word-Word & 0.759 \\
    \midrule
    \multirow{7}{*}{SBERTScore} & Word-Word & 0.767\\
     & Sent-Sent & \textbf{0.779} \\
     & Doc-Sent & 0.576 \\
     & Sent-Doc & 0.746 \\
     & Doc-Doc & 0.684 \\
     & Mean-Sent & 0.602 \\
     & Sent-Mean & 0.565 \\
     & Mean-Mean & 0.512 \\
    \bottomrule
    \end{tabular}
    \caption{Balanced accuracy with different text granularities as input. The highest balanced accuracy is highlighted in \textbf{bold}, which is significantly higher than the second best result with $p<0.05$.}
    \label{tab:granularity}
\end{table}
For BERTScore, we only test word level embeddings since it has been reported that BERT does not perform well in representing higher level text embeddings \cite{reimers-2019-sbert}. 
For SBERTScore, we additionally test word level input to better understand the contribution of granularity to the improvement.

SBERTScore on sentence-sentence level achieves the highest score in Table \ref{tab:granularity}. It also outperforms BERTScore on the same word-word level similarity, indicating that the improvement is brought by both the architecture and the appropriate text granularity.
For document-level, the performance drops greatly when it is applied on the source document, as 45.76\% of the source documents are trunctated.
Inputting the summary at document level has a much smaller effect as the summary length is usually much shorter than the length limit. 
Segmenting the source documents at the right granularity can avoid the information loss brought by the length limit while producing more suitable embeddings for judging factuality.

A simplification to SBERTScore is to compute the mean sentence embedding for an input document, avoiding the need to search for the maximum similarity while still processing sentences individually with SBERT. 
In Table \ref{tab:granularity}, we observe that averaging either source or summary will lead to worse balanced accuracy, which justifies the sentence granularity proposed in Section \ref{sec:sbertscore}.

\subsection{Case Study: Negation}
BERTScore is reported to struggle at handling negation accurately \cite{leiter-2022-towards}. We conduct a case study to investigate the performance of SBERTScore when processing negation. Consider the four examples sentences below:
\begin{itemize}
    \item[$S_1$] I like rainy days because they make me feel relaxed
    \item[$S_2$] I don't like rainy days because they don't make me feel relaxed.
    \item[$S_3$] I enjoy rainy days because they make me feel calm.
    \item[$S_4$] I enjoy listening to music at rainy days.
\end{itemize}
Table \ref{tab:case} shows the BERTScores and SBERTScores obtained by comparing the given sentence pairs. 
BERTScore fails to identify the negation in $S_2$ and assigns a high score despite its inconsistency with $S_1$. 
SBERTScore does better since it works on the sentence-level where negation could have a larger influence. 
However, the comparison between SBERTScores of $\langle S_1, S_2 \rangle$ and $\langle S_1, S_4 \rangle$ indicates that it is not sensitive enough to distinguish between negation and neutral expressions.
$\langle S_1, S_4 \rangle$ do not contradict one another, so should receive a higher score, yet both pairs have very similar SBERTScores. 
Future research is therefore required into handling negation.

\begin{table}[htbp]
    \centering
    \small
    \begin{tabular}{lccc}
        \toprule
        Metric & $\langle S_1, S_2 \rangle$ & $\langle S_1, S_3 \rangle$ & $\langle S_1, S_4 \rangle$ \\
        \midrule
        BERTScore & 0.984 & 0.988 & 0.915 \\
        SBERTScore & 0.720 & 0.975 & 0.701\\
        \bottomrule
    \end{tabular}
    \caption{BERTScore and SBERTScore of example sentence pairs.}
    \label{tab:case}
\end{table}

\subsection{Benchmark Comparison with NLI and QA-based Methods}

\begin{table*}[htbp]
    \centering
    \small
    \begin{tabular}{lcccccc}
    \toprule
     \multirow{2}{*}{Metric} & \multicolumn{2}{c}{CNNDM} & \multicolumn{2}{c}{XSum} & \multicolumn{2}{c}{Overall} \\
    \cmidrule(lr){2-3}\cmidrule(lr){4-5}\cmidrule(lr){6-7}
     & Banlanced Acc. & ROC-AUC & Banlanced Acc. & ROC-AUC & Banlanced Acc. & ROC-AUC \\
    \midrule
     QAFactEval & \textbf{0.757} & \textbf{0.823} & \textbf{0.705} & \textbf{0.773} & \textbf{0.817} & \textbf{0.883} \\
     QuestEval & 0.670 & 0.736 & 0.665 & 0.711 & 0.758 & 0.843 \\
     DAE & 0.696 & 0.747 & - & - & - & - \\
     SummaC$_{Conv}$ & 0.737 & 0.796 & 0.604 & 0.654 & 0.789 & 0.857 \\
     \midrule
     SummaC$_{ZS}$ & 0.686 & 0.759 & 0.577 & 0.607 & \textbf{0.782} & 0.836 \\
     BERTScore & 0.692 & 0.767 & \textbf{0.695} & \textbf{0.738} & 0.759 & 0.832 \\
     SBERTScore & \textbf{0.720} & \textbf{0.804} & 0.605 & 0.653 & 0.779 & \textbf{0.851} \\
    \bottomrule
    \end{tabular}
    \caption{Balanced accuracy and ROC-AUC of different metrics on each dataset split. Metrics in the top require training while the bottom ones are zero-shot. The best results of each column on the two sections are \textbf{highlighted} and are significantly better than the next best one in their section with $p<0.05$. Following the setting of \citet{tang-2023-understanding}, we remove the results of DAE for a fair comparison as it is trained on the annotated validation set of XSum.}
    \label{tab:main}
\end{table*}

\begin{table*}[htbp]
    \centering
    \small
    \begin{tabular}{lccccccccc}
    \toprule
    \multirow{2}{*}{Metric} & \multicolumn{9}{c}{Dataset} \\
    \cmidrule(lr){2-10}
     & XSF & Polytope & FactCC & SummEval & FRANK & QAGS & CLIFF & Goyal' 21 & XENT  \\
    \midrule
     QAFactEval & 0.604 & \underline{\textbf{0.827}} & 0.843 & \underline{\textbf{0.830}} & \underline{\textbf{0.729}} & \underline{\textbf{0.692}} & 0.703 & 0.754 & 0.613 \\
     QuestEval & 0.605 & 0.708 & 0.655 & 0.713 & 0.567 & 0.607 & 0.691 & \underline{\textbf{0.797}} & 0.601 \\
     DAE & - & 0.782	& 0.704	& 0.716	& 0.695 & 0.586 & 0.734 & - & - \\
     SummaC$_{Conv}$ & \underline{\textbf{0.655}} & 0.744 & \underline{\textbf{0.891}} & 0.793 & 0.655 & 0.629 & \textbf{0.744} & 0.552 & \underline{\textbf{0.668}} \\
     \midrule
     SummaC$_{ZS}$ & 0.549 & \textbf{0.786} & \underline{\textbf{0.835}} & 0.781 & 0.672 & \underline{\textbf{0.673}} & 0.700 & 0.466 & 0.490\\
     BERTScore & 0.527 & 0.779 & 0.632 & 0.759 & \textbf{0.676} & 0.586 & \underline{\textbf{0.724}} & \underline{\textbf{0.657}} & \underline{\textbf{0.601}}\\
     SBERTScore & \underline{\textbf{0.608}} & 0.772 & 0.754 & \underline{\textbf{0.827}} & 0.655 & 0.596 & 0.701 & 0.605 & 0.581\\
    \bottomrule
    \end{tabular}
    \caption{Balanced accuracy of different metrics on each dataset. Metrics in the top require training while the bottom ones are zero-shot. The best results of each column in the two sections are \textbf{highlighted}. \underline{Underline} indicates the result is significantly better than the second best one in the same section with $p<0.05$. We only report the DAE's results on CNNDM and remove the results on the part of the dataset that only contains XSum data.}
    \label{tab:detail_main}
\end{table*}

In Table \ref{tab:main}, we combine the data from the same origin to compute ROC-AUC and set a single threshold for them to compute the balanced accuracy. The last two columns are the results obtained after mixing all data.
QAFactEval outperforms other metrics on all splits of the dataset.
Other metrics are competitive with each other as they all have advantageous and disadvantageous datasets.
Along with the detailed results in Table \ref{tab:detail_main}, we find that metric performance varies across different datasets, suggesting that choosing a suitable metric will, in practice, depend on the dataset. 

Given suitable comparison text, BERTScore is actually much better than previous studies~\cite{fabbri-2021-qafacteval,pagnoni-2021-understanding,durmus-2020-feqa}, it outperforms all zero-shot metrics and two other trained metrics on the XSum split.
SBERTScore outperforms SummaC$_{ZS}$ on all dataset splits except being slightly lower on overall balanced accuracy.
In terms of ROC-AUC, it achieves the second highest on CNNDM and is third highest on the whole dataset, demonstrating better factuality classification ability than some recent metrics that use either NLI or QA paradigms, especially comparing to SummaC$_{ZS}$ that also uses the zero-shot setting. 
Their performance indicates that similarity-based metrics are still promising and competitive with recent factuality metrics.
SBERTScore outperforms BERTScore on CNNDM and overall scores but underperforms on XSum. 
We speculate that is because most XSum summaries are a single sentence, which prevents our proposed metric from averaging scores over sentences and leads to degeneration. 
Some evidence for this is that SummaC$_{ZS}$, which averages the maximum scores in each column of the score matrix in the same way as our metric, also underperforms on XSum. 
However, both Summac$_{Conv}$ and BERTScore, as comparable alternatives to these two metrics, still average scores from several comparisons, thus having better performance.

\subsection{Error Analysis and Metric Combination}
Previous studies \cite{pagnoni-2021-understanding,tang-2023-understanding} point out that different metrics can be sensitive to different errors, inspiring us to look into the possibility of combining different metrics. We first investigate the error type sensitivity of BERTScore and SBERTScore, following the coarse error type taxonomy in \cite{tang-2023-understanding}.
Errors are classified from two perspectives. Errors made up by text pieces that appear in the source document are noted as $Intrinsic$, otherwise $Extrinsic$. 
The error attributes are furthered classified as either $Noun Phrase$ or $Predicate$. 
All errors from XSF \citep{maynez-2020-faithfulness}, FRANK \citep{pagnoni-2021-understanding}, Goyal 21' \citep{goyal-2021-annotating}, and CLIFF \citep{cao-2021-cliff} are annotated with a subset of $\{Intrinsic, Extrinsic\} \times \{Noun Phrase, Predicate\}$. 
For summaries from XSum, they have two special additional error types, $\{Intrinsic Sentence, Extrinsic Sentence\}$, if the whole sentence is inconsistent. 
The error analysis investigates each metric's recall on detecting certain type of errors, as well as correct summaries, as shown in Table \ref{tab:errors}.

The results in Table \ref{tab:errors} demonstrate that metrics have different strengths.
Benefiting from the properties of similarity, BERTScore and SBERTScore perform better on extrinsic than intrinsic errors for the same attribute type. 
Compared to the recall of errors, the most impressive ability of SBERTScore is to identify correct summaries. 
It significantly outperforms all the other metrics on CNNDM, and comes only after SummaC$_{ZS}$ on XSum. 

\begin{table*}[htbp]
    \centering
    \small
    \begin{tabular}{l ccccc cccccccc@{}}
    \toprule
       \multirow{3}{*}{Metric}  & \multicolumn{5}{c}{CNNDM} & \multicolumn{7}{c}{Xsum} & \\
    \cmidrule(lr){2-6} \cmidrule(lr){7-13}
         & \multicolumn{2}{c}{Intrinsic} & \multicolumn{2}{c}{Extrinsic} & \multirow{2}{*}{Correct} & \multicolumn{3}{c}{Intrinsic} & \multicolumn{3}{c}{Extrinsic} &  \multirow{2}{*}{Correct} \\
    \cmidrule(lr){2-3} \cmidrule(lr){4-5} \cmidrule(lr){7-9} \cmidrule(lr){10-12}
         & NP. & P. & NP & P. &  & NP. & P. & Sent. & NP & P. & Sent. & \\
    \midrule
         QAFactEval & 0.546 & 0.509 & 0.791 & 0.633 & 0.401 & \underline{\textbf{0.671}} & \underline{\textbf{0.720}} & 0.882 & 0.532 & 0.631 & 0.808 & 0.304 \\
         QuestEval & \textbf{0.695} & 0.582 & 0.777 & \underline{\textbf{0.742}} & 0.309 & 0.493 & 0.553 & \underline{\textbf{0.941}} & 0.520 & \underline{\textbf{0.644}} & \underline{\textbf{0.849}} & \underline{\textbf{0.387}} \\
         DAE & 0.575 & 0.509 & 0.668 & 0.609 & \underline{\textbf{0.436}} & - & - & - & - & - & - & - \\
         SummaC$_{Conv}$ & 0.684 & \underline{\textbf{0.782}} & \underline{\textbf{0.841}} & 0.711 & 0.287 & 0.551 & 0.629 & 0.294 & \underline{\textbf{0.640}} & 0.619 & 0.715 & 0.371 \\
        \midrule
         SummaC$_{ZS}$ & 0.632 & \underline{\textbf{0.745}} & \underline{\textbf{0.800}} & 0.711 & 0.314 & \underline{\textbf{0.676}} & \underline{\textbf{0.652}} & 0.824 & 0.569 & 0.589 & 0.523 & \underline{\textbf{0.418}} \\
         BERTScore & \underline{\textbf{0.661}} & 0.636 & 0.741 & \textbf{0.719} & 0.342 & 0.538 & 0.621 & \underline{\textbf{0.882}} & \underline{\textbf{0.597}} & 0.631 & 0.782 & 0.375 \\
         SBERTScore & 0.454 & 0.436 & 0.586 & 0.563 & \underline{\textbf{0.522}} & 0.498 & 0.644 & 0.706 & 0.532 & \underline{\textbf{0.661}} & \underline{\textbf{0.808}} & 0.397 \\
    \bottomrule
    \end{tabular}
    \caption{Recall of each metric on different types of errors, as well as correct summaries. Metrics in the top require training while the bottom ones are zero-shot. The best results of each column in the two sections are \textbf{highlighted}. \underline{Underline} indicates the result is significantly better than the second best  in the same section with $p<0.05$. We remove the results of DAE for a fair comparison as it is trained on the annotated validation set of XSum.}
    \label{tab:errors}
\end{table*}

Furthermore, we investigate the agreement among different metrics on the benchmark to find out whether they can be complementary to each other. 
The Kohen's $\kappa$ scores in Appendix \ref{sec:agree} show weak agreement ($<0.45$) among the metrics. 
Considering that these metrics have similar balanced accuracy, it suggests that a combination of comparison approaches could be more effective than relying on a single metric.
We simply test this idea by combining pairs of distinct evaluation metrics using logical \textit{AND} and \textit{OR}. 

\begin{figure}[htbp]
    \centering
    \includegraphics[width=\linewidth]{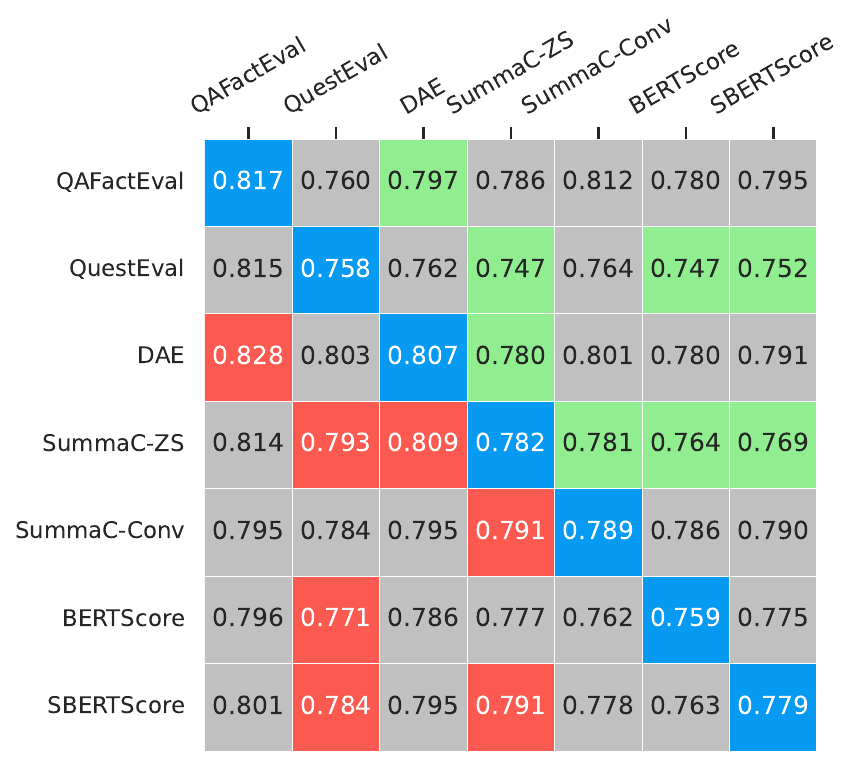}
    \caption{Average balanced accuracy of combined metrics on the benchmark. The diagonal is the balanced accuracy of the original evaluation metric (highlighted in blue). The upper triangular matrix is the balanced accuracy of joint metrics using \textit{OR} and the lower triangular matrix is based on \textit{AND}. Red blocks highlight the balanced accuracy that is improved over two original metrics, and green blocks highlight those are lower than both original metrics. All improvements and declines are statistically significant with $p<0.05$.}
    \label{fig:combination}
\end{figure}

The joint balanced accuracy of each combination is shown in Figure \ref{fig:combination}. The lower triangular matrix indicates that logical \textit{AND} can improve the balanced accuracy, while the upper triangular matrix suggest opposite to logical \textit{OR}. 
Since \textit{OR} marks a summary as consistent if either of the base metrics classifies it as such, it demonstrates that individual factuality metrics may suffer from false positives. 
Logical \textit{AND} introduces a double-checking mechanism, which raises the accuracy by mitigating the false consistent rate and improving the true inconsistent rate. 
We show a combination example using SBERTScore and QuestEval in Appendix \ref{sec:combination_example}. 

\section{Conclusion}
In this paper, we investigated the suitable settings for similarity-based factuality evaluation metrics and propose a new sentence-sentence level metric, SBERTScore. 
We show that, given source documents as input, similarity-based evaluation metrics computed on sentence-sentence level are competitive with more complex NLI and QA-based factuality-oriented metrics, and do not require a supervised learning step on the target domain.
Also, our proposed metric better aligns with human binary annotations than the widely-used BERTScore on CNNDM subset and overall dataset on the benchmark. 
It outperforms the weaker baselines using NLI and QA-based paradigms and achieves competitive balanced accuracy with the strongest fine-tuned NLI-based metric.
Therefore, we conclude that zero-shot similarity-based metrics are a promising approach.
We analyse the advantages of our proposed metric in detecting correct summaries, investigate the agreement among different metrics, and find that similarity-based metrics make different errors to QA and NLI-based metrics. 
Building on this, we show that integrating metrics by logical \textit{AND} can improve balanced accuracy on benchmark datasets. 
Furthermore, we illustrate a limitation of similarity-based metrics when processing negation and highly similar but neutral input text, which suggests a direction for future research.

\newpage

\section*{Limitations}
The proposed metric in this paper shows competitive performance comparing to strong factuality metrics and can be used out-of-box. 
However, our proposed metric is based on similarity, which is insufficient for precisely detecting factual errors, because high similarity cannot guarantee factual consistency.
Our case study shows that although SBERTScore can handle negation better, it still cannot distinguish highly similar sentences that are actually neutral to each other.
Our investigation into metric combination represents only an initial step.
The results of error analysis and inter-metric agreement suggest that designing more sophisticated methods for combining these metrics may be a promising way to make progress in future work.
We note that our experiments are limited to English news datasets, and suggest that further investigation is needed to develop and test factuality approaches for other languages and text domains. 

\section*{Acknowledgments}

\bibliography{acl_latex}

\appendix

\section{Metric Processing Speed}
\label{sec:speed}

We randomly sampled 1000 pieces of data from the benchmark and ran QuestEval, SummaC$_{\{ZS,Conv\}}$, BERTScore and SBERTScore on them. We didn't test DAE and QAFactEval as their dependencies are not compatible with our GPU. The runtime of each metric to processing 1000 pieces of data is presented in Table \ref{tab:speed}.

\begin{table}[htbp]
  \centering
  \small
  \begin{tabular}{lc}
    \toprule
    \textbf{Metric} & \textbf{Time (s)} \\
    \hline
    QuestEval & 1914 \\
    SummaC$_{ZS}$ & 207 \\ 
    SummaC$_{Conv}$ & 233 \\
    BERTScore & \textbf{36} \\
    SBERTScore & 67 \\
    \bottomrule
  \end{tabular}
  \caption{The total time needed for each metric to process the 1000 pieces of samples. The fastest metric is \textbf{highlighted}.}
  \label{tab:speed}
\end{table}

\section{Inter-Metric Agreement}
\label{sec:agree}
\begin{figure}[htbp]
    \centering
    \includegraphics[width=\linewidth]{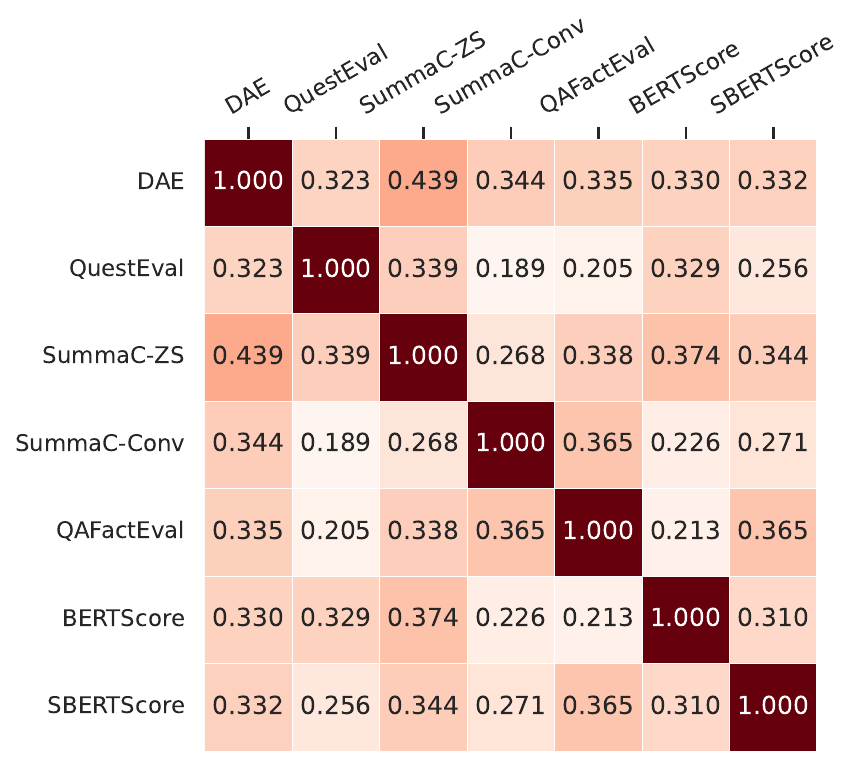}
    \caption{Cohen's $\kappa$ agreement score among different metrics on the benchmark dataset. The higher agreement is in deeper red.}
    \label{fig:agreement}
\end{figure}

We compute Cohen's $\kappa$ among all metrics using their binary predictions on the benchmark. Figure \ref{fig:agreement} shows the agreement between the metrics.

\section{Example of Metric Combination}
\label{sec:combination_example}
We use SBERTScore and QuestEval as a combination example where two metrics work in a complementary way to correct the false judgement.
\begin{table}[htbp]
    \centering
    \small
    \begin{tabular}{|p{0.14\columnwidth}|p{0.75\columnwidth}|}
    \hline
        Source & Sidwell, 34, has made 32 Championship appearances this season to help the Seagulls achieve promotion to the top flight for the first time in 34 years. With his contract due to expire at the end of the campaign, the midfielder is now hoping to sign a new deal. "I want to be a part of it next year because I know we can stay in the Premier League," he said, "...it can be done and we can enjoy the summer." \\
        \hline
        Summary & steven sidwell says he wants to stay at brighton until the end of the season. \\
    \hline
    \end{tabular}
    \caption{An example extracted from the benchmark dataset.}
    \label{tab:and_case}
\end{table}
Table \ref{tab:and_case} shows a story extracted from the benchmark dataset. The source and summary pair have an SBERTScore of $0.610$, which marks it as factually inconsistent.
QuestEval gives $0.426$ with consistent judgement, probably because the major noun chunks in the summary are covered by the source document, but they are actually used incorrectly.
Logical \textit{AND} takes two labels into consideration and decides the final prediction as \emph{inconsistent} which corrects the false positive prediction from QuestEval.

The confusion matrices of the base metrics and the  \textit{AND} combination, shown below in (Table \ref{tab:fp}), support our inference that combination can mitigate false consistent (false positive, FP) and improve true inconsistent (TP) rates.

\begin{table}[htbp]
    \centering
    \small
    \begin{tabular}{ccccc}
    \toprule
     Metric & TP & TN & FP & FN \\
    \midrule
    SBERTScore & 0.444 & 0.332 & 0.084 & 0.141 \\
    QuestEval & 0.511 & 0.266 & 0.150 & 0.074 \\
    Combined & 0.418 & 0.355 & 0.061 & 0.166\\
    \bottomrule
    \end{tabular}
    \caption{Confusion matrices of different metrics and their combined metric on the benchmark.}
    \label{tab:fp}
\end{table}

\end{document}